\PassOptionsToPackage{table}{xcolor}
\documentclass[sigconf]{acmart}

\AtBeginDocument{%
  }

\setcopyright{none}
\copyrightyear{2025}
\acmYear{2025}

\acmConference[Prompt Optimization KDD '25]{First International KDD Workshop on Prompt Optimization, 2025}{August 4, 2025}{Toronto, ON, Canada}

\usepackage[table]{xcolor}
\usepackage{tabularx}
\usepackage{booktabs}
\usepackage{array}
\usepackage{multirow}
\usepackage{mdframed}
\usepackage{algorithm}
\usepackage{algorithmic}
\usepackage{tcolorbox}

\definecolor{col1}{RGB}{230,242,255}
\definecolor{col2}{RGB}{255,242,230}
\definecolor{col3}{RGB}{242,255,230}
\definecolor{col4}{RGB}{255,230,242}

\begin{document}

%%
%% Title
\title{Plan-and-Write: Structure-Guided Length Control for LLMs without Model Retraining}

%%
%% Authors
\author{Adewale Akinfaderin}
\email{akinfaa@amazon.com}
\affiliation{%
  \institution{Amazon Web Services}
  \city{Seattle}
  \state{WA}
  \country{USA}
}

\author{Shreyas Subramanian}
\email{subshrey@amazon.com}
\affiliation{%
  \institution{Amazon Web Services}
  \city{Seattle}
  \state{WA}
  \country{USA}
}

\author{Akarsha Sehwag}
\email{akshseh@amazon.com}
\affiliation{%
  \institution{Amazon Web Services}
  \city{Seattle}
  \state{WA}
  \country{USA}
}

%%
%% Short author list
\renewcommand{\shortauthors}{Akinfaderin et al.}

%%
%% Abstract
\begin{abstract}
Length control in Large Language Models (LLMs) is a crucial but under-addressed challenge, with applications ranging from voice interfaces requiring concise responses to research summaries needing comprehensive outputs. Current approaches to length control, including Regularized DPO, Length-Instruction Fine-Tuning, and tool-augmented methods, typically require expensive model retraining or complex inference-time tooling. This paper presents a prompt engineering methodology that enables precise length control without model retraining. Our structure-guided approach implements deliberate planning and word counting mechanisms within the prompt, encouraging the model to carefully track and adhere to specified length constraints. Comprehensive evaluations across six state-of-the-art LLMs demonstrate that our method significantly improves length fidelity for several models compared to standard prompting when applied to document summarization tasks, particularly for shorter-to-medium length constraints. The proposed technique shows varying benefits across different model architectures, with some models demonstrating up to 37.6\% improvement in length adherence. Quality evaluations further reveal that our approach maintains or enhances overall output quality compared to standard prompting techniques. Our approach provides an immediately deployable solution for applications requiring precise length control, particularly valuable for production environments where model retraining is impractical or cost-prohibitive.
\end{abstract}

%%
%% CCS Concepts
\ccsdesc[500]{Computing methodologies~Natural language processing}
\ccsdesc[300]{Computing methodologies~Machine learning}
\ccsdesc{Human-centered computing~Natural language interfaces}
\ccsdesc[100]{Information systems~Language models}

%%
%% Keywords
\keywords{Prompt Engineering, Large Language Models, Length Control \& Fidelity}

%%
%% Build title
\maketitle

%%
%% INTRODUCTION
\section{Introduction}

Length control is a fundamental yet frequently overlooked aspect of language model capabilities. As Large Language Models (LLMs) become increasingly integrated into real-world applications, the ability to precisely control response length emerges as a critical requirement. Different use cases demand different response lengths: voice interfaces require concise answers, research summaries need comprehensive detail, mobile applications have screen space constraints, and documentation systems must provide varying levels of information density \cite{yuan2024following}. Despite their remarkable progress in reasoning and knowledge tasks, even state-of-the-art LLMs struggle with this basic aspect of communication, often violating explicit length constraints in numerous cases \cite{yuan2024following, dubois2024length}.

The inability to control output length creates significant challenges for production deployments. Unpredictable response lengths lead to poor user experiences, increased computational costs, and reduced system reliability. Moreover, many LLMs exhibit a systematic bias toward verbosity, a phenomenon that compromises both their efficiency and utility \cite{dubois2024length, nayab2024concise}. While several methods have been proposed to address this issue, these approaches typically require expensive model retraining, complex fine-tuning procedures, or resource-intensive inference-time mechanisms \cite{gu2024length, yuan2024following}. These limitations make them impractical for many production environments, particularly those utilizing black-box LLMs where model parameters are inaccessible \cite{gu2024length}.

Our work introduces ``Plan-and-Write,'' a prompt engineering methodology that enables precise length control without model retraining. Unlike existing approaches, our method operates entirely within the prompt space, making it immediately deployable across any LLM without additional training or tooling requirements. The core innovation of our approach lies in a structure-guided planning technique that encourages models to strategically allocate their word budget across content sections. By incorporating explicit planning and word-counting mechanisms within carefully designed prompts, we guide LLMs to maintain awareness of their output length while preserving response quality.

%%
%% Example table
\begin{table*}[h!]
\caption{Example length-controlled document summarization using different prompting strategies. The vanilla approach often fails to meet exact target lengths, while our thinking approach demonstrates precise word count control.}
\label{tab:examples}
\small
\begin{tabularx}{\textwidth}{>{\columncolor{col1}\raggedright}p{6cm}>{\columncolor{col2}}c>{\columncolor{col3}}X>{\columncolor{col4}}c}
\toprule
\textbf{Instruction} & \textbf{Target} & \textbf{Generation} & \textbf{Length} \\
\midrule
\parbox[t]{6cm}{\texttt{Summarize this document in exactly \{target\_words\} words.}}
& 20 & Amazon had a strong year in 2023, with revenue, operating income, and free cash flow growth. The company is investing in primitives and generative AI capabilities. & 26 \\
\parbox[t]{6cm}{}
& 50 & Amazon had a strong year in 2023, with revenue growth, improved profitability, and continued customer experience enhancements. The company is focused on building foundational ``primitive'' services to rapidly innovate and empower both internal and external builders, with a particular emphasis on generative AI capabilities. & 46 \\
\parbox[t]{6cm}{}
& 100 & Amazon's 2023 annual letter to shareholders highlights the company's strong financial performance, with 12\% revenue growth and a dramatic improvement in operating income and free cash flow. The letter emphasizes Amazon's focus on customer experience, with enhancements in selection, pricing, and delivery speed. It also discusses the company's progress in Advertising, AWS, and newer business investments like Prime Video and Project Kuiper. The letter emphasizes Amazon's ``primitives'' approach, which involves building foundational services to enable rapid innovation. It also outlines the company's vision for Generative AI and its potential to transform various Amazon businesses. Overall, the letter conveys Amazon's optimism and conviction in its long-term growth and innovation potential. & 118 \\
\midrule
\parbox[t]{6cm}{\texttt{YOUR ONLY TASK: Summarize this document in EXACTLY \{target\_words\} WORDS.}}
& 20 & Amazon's strong financial results, customer experience, and focus on building primitive services to empower builders and innovation across businesses. & 20 \\
\parbox[t]{6cm}{\texttt{Count each word as you write.}}
& 50 & Amazon saw strong growth in 2023, with revenue, operating income, and free cash flow improving significantly. The company is investing in customer experience, logistics, and advertising, while also making progress in AWS, Prime Video, and new initiatives like Generative AI and Project Kuiper. Amazon remains focused on long-term value creation. & 50 \\
\parbox[t]{6cm}{\texttt{Rewrite in paragraph form.}}
& 100 & Amazon saw strong growth in 2023, with revenue increasing 12\% to \$575 billion. Operating income and free cash flow also improved significantly. The company attributed these results to its focus on customer experience, including expanded selection, competitive pricing, and faster delivery. Amazon continued investing in key initiatives like AWS, Prime Video, and Project Kuiper. The letter discusses the company's ``primitives'' approach to building flexible, reusable services that enable rapid innovation. It also highlights Amazon's focus on generative AI as a transformative technology. Overall, the letter conveys the company's enthusiasm and optimism for the future. & 99 \\
\bottomrule
\end{tabularx}
\end{table*}

We evaluate our approach across seven state-of-the-art LLMs on document summarization tasks with varying target length requirements. Our results demonstrate that the Plan-and-Write methodology significantly improves length fidelity compared to standard prompting techniques, with minimal impact on response quality. We identify patterns in model behavior that suggest advanced LLMs possess untapped capabilities for length control that can be effectively activated through structured prompting. See Table~\ref{tab:examples} for examples of length-controlled generations using both standard prompting and our structure-guided approach. Our evaluation shows that our method achieves substantially higher precision in meeting exact word count targets while maintaining or enhancing the quality and relevance of model responses.

%%
%% RELATED WORK
\section{Related Work}

Length control in LLMs has emerged as a critical challenge as these models become increasingly integrated into real-world applications with specific output constraints. Research on text generation control dates back to earlier work by Holtzman et al. \cite{holtzman2018learning}, who explored cooperative discriminators to guide language generation. However, as noted in recent benchmarks, modern LLMs still struggle with this fundamental aspect of communication, often violating explicit length constraints in over 50\% of cases \cite{yuan2024following}. Current research addressing this challenge has primarily explored three categories of approaches: model fine-tuning, inference-time modifications, and prompt engineering techniques.

Several works have focused on modifying model parameters through specialized fine-tuning procedures. Yuan et al. \cite{yuan2024following} proposed Length-Instruction Fine-Tuning (LIFT), which augments preference datasets with examples where responses violating length constraints automatically lose in preference pairs. Zhou et al. \cite{zhou2024t} introduced token-level reward regularization (T-REG) to improve preference optimization, which could benefit length control among other aspects. Park et al. \cite{park2024regularized} developed Regularized Direct Preference Optimization to address length exploitation in RLHF by adding a principled regularization term. Dubois et al. \cite{dubois2024length} identified systematic biases toward verbosity in LLMs and proposed a causal inference framework to debias evaluation metrics. These approaches show promising results but require expensive retraining procedures and access to model weights, making them impractical for many deployment scenarios involving black-box LLMs.

Inference-time methods attempt to control output length without modifying model parameters. Gu et al. \cite{gu2024length} developed an iterative sampling framework based on the Metropolis-Hastings algorithm that treats length control as sampling from a target distribution. Their approach requires multiple inference passes. Nayab et al. \cite{nayab2024concise} explored the relationship between output verbosity and reasoning quality, proposing Constrained Chain-of-Thought prompting to explicitly limit reasoning length. While these methods offer more flexibility than fine-tuning approaches, they often introduce additional computational overhead during inference.

More closely aligned with our approach, prompt engineering techniques attempt to achieve length control through careful instruction design without modifying models or using complex inference procedures. Zhang et al. \cite{zhang2023rest} demonstrated that structured prompting can guide LLMs to better understand and follow formatting constraints in data representation tasks. Lyu et al. \cite{lyu2022z} explored zero-shot in-context learning with pseudo-demonstrations, showing how carefully constructed prompts can guide model behavior without parameter updates. Bai et al. \cite{bai2022constitutional} introduced constitutional AI techniques that improve instruction following in general, which indirectly benefits length constraint adherence. Wang et al. \cite{wang2022learning} explored a primal-dual approach for controlled question generation that incorporates specific constraints during generation. Despite these advances, there remains a significant gap between theoretical approaches and practical, deployment-ready solutions for precise length control. Most current methods either sacrifice response quality or work only for specific types of constraints. Our Plan-and-Write methodology addresses these limitations through a prompt engineering approach that enables precise word-count control without model retraining or additional inference overhead, while maintaining response quality.

%%
%% METHODOLOGY
\section{Methodology}

\subsection{Overview}

We introduce Plan-and-Write, a prompt engineering methodology for precise length control in large language models. The core innovation of our approach is the decomposition of the generation process into two distinct phases:
\begin{enumerate}
    \item a planning phase where the model explicitly counts words as it drafts content, and
    \item a verification phase where the model reconstructs the content into a coherent output of exactly the specified length.
\end{enumerate}

This structure-guided approach leverages the LLM's inherent capabilities for metacognition and self-monitoring without requiring any model parameter modifications or additional inference passes.

Unlike traditional approaches that simply include a length constraint in the instruction (e.g., ``Summarize in X words''), our method guides the model through a deliberate process of word counting and budget allocation. This creates what we term ``length awareness''---an explicit tracking mechanism that helps the model maintain precise control over its output length while preserving content quality. The approach is model-agnostic and can be applied to any LLM capable of following multi-step instructions.

\subsection{Framework}

We formally define the length-controlled generation problem as follows: Given a document $D$, a target word count $t$, and an LLM $M$, find a summary $S$ that maximizes the semantic relevance to $D$ while strictly adhering to the target length:

\begin{equation}
S = \arg\max_{s} \text{Quality}(s, D) \quad \text{subject to} \quad |s| = t
\end{equation}

where $|s|$ represents the word count of summary $s$, and Quality$(s, D)$ measures semantic similarity and relevance between the summary and the original document.

Traditional approaches attempt to solve this directly with a simple constraint in the prompt:

\begin{equation}
S_{\text{vanilla}} = M(D, P_{\text{vanilla}}(t))
\end{equation}

where $P_{\text{vanilla}}(t)$ represents a prompt instructing the model to generate a summary of length $t$.

Our Plan-and-Write approach decomposes this into two phases:

\begin{equation}
S_{\text{draft}} = f_{\text{planning}}(M, D, t)
\end{equation}

\begin{equation}
S_{\text{final}} = f_{\text{verify}}(S_{\text{draft}}, t)
\end{equation}

This decomposition enables the model to first plan a response with explicit word counting, followed by a verification step that ensures the exact target length is met. We measure the effectiveness of our approach using length fidelity error:

\begin{equation}
E = ||~|S| - t~||
\end{equation}

with the goal of minimizing $E$ to 0 while maintaining high quality output.

\paragraph{Phase 1: Planning with Explicit Word Counting}
The model generates content while numbering each word sequentially, creating ``length awareness'' throughout the generation process.

\paragraph{Phase 2: Verification and Coherence}
The model reformats the same content into a coherent paragraph while maintaining the exact word count, ensuring both precision in length and quality in expression.

The final response is extracted from between the designated tags, providing a clean output that meets the exact target length. This approach requires no model retraining or additional inference passes, making it immediately deployable with any LLM capable of following multi-step instructions.

\subsection{Justification}

Our approach is based on three simple but effective principles that explain why explicit word counting improves length control in LLMs:

\paragraph{Explicit Monitoring}
By instructing the model to count words as it generates, we create a simple tracking mechanism that makes the length constraint concrete rather than abstract. This transforms the vague instruction ``write exactly $t$ words'' into a procedural task with clear progress indicators.

\paragraph{Two-Phase Structure}
Separating content generation from final formatting allows the model to first focus on meeting the word count exactly, then focus on making the text coherent and fluent. This division of the task reduces the cognitive burden on the model by addressing one constraint at a time.

\paragraph{Leveraging Existing Capabilities}
Rather than requiring new model capabilities through fine-tuning, our method simply provides a process that helps the model apply its existing abilities more effectively. Modern LLMs can count and follow instructions---our prompt structure just guides them to apply these skills to solve the length control problem.

This straightforward approach explains why Plan-and-Write achieves better length fidelity than traditional prompting methods. By providing a clear process rather than just stating a constraint, we help the model systematically achieve exact word counts while maintaining response quality.

%%
%% EXPERIMENTS
\section{Experiments}

We designed a comprehensive evaluation framework to assess the effectiveness of our Plan-and-Write methodology across multiple dimensions, including model capabilities, task types, and target lengths. Our experiments focused on comparing the length fidelity of our structure-guided approach against standard prompting techniques while maintaining output quality.

\subsection{Experimental Setup}

\subsubsection{Models}
We evaluated our approach on six state-of-the-art large language models representing diverse training methodologies and architectural designs: Claude 3 Haiku, Claude 3.5 Haiku, Claude 3.5 Sonnet, Claude 3.7 Sonnet, Mistral Large, and Meta's Llama 3.1 70B. These models span different parameter sizes and capabilities, allowing us to assess the generalizability of our approach across the current landscape of LLMs.

\subsubsection{Tasks and Target Lengths}
We conducted experiments on document summarization, where models were provided with a PDF document (Amazon's 2023 Shareholder Letter) and instructed to create summaries of varying specified lengths. To evaluate length control across different magnitudes, we tested eight target word counts: 20, 50, 100, 200, 500, 1000, 2000, and 5000 words. This range enabled us to observe model performance on both extremely concise outputs and more extensive generations.

\subsubsection{Evaluation Metrics}
Our primary metric was Mean Absolute Percentage Deviation (MAPD), which measures the percentage error between generated and target word counts:

\begin{equation}
\text{MAPD} = \frac{|\text{generated\_words} - \text{target\_words}|}{\text{target\_words}}
\end{equation}

MAPD provides a normalized measure of length fidelity independent of target size, enabling fair comparison across different length targets.

\subsection{Implementation Details}

\subsubsection{Document Processing}
For the summarization task, we provided the document as a PDF file through the models' document understanding capabilities. This allowed us to test length control in a realistic setting where models must process and condense complex multi-page documents.

\subsubsection{Prompt Variants}
We designed and evaluated four distinct prompting strategies to systematically test different implementations of length control. The prompts are provided below:

\begin{enumerate}
    \item \textbf{Vanilla V1}: A straightforward instruction to summarize in the target word count.

\begin{tcolorbox}[
  title={\textbf{Vanilla V1 Prompt}},
  colback=blue!10,
  colframe=blue!50,
  coltitle=black,
  boxrule=1.5pt,
  arc=3mm,
  boxsep=5pt,
  left=5pt,
  right=5pt
]
Summarize this document into exactly \{target\_words\} words.
\end{tcolorbox}

\item \textbf{Vanilla V2}: A variation of the standard prompt using different phrasing.

\begin{tcolorbox}[
  title={\textbf{Vanilla V2 Prompt}},
  colback=green!10,
  colframe=green!50,
  coltitle=black,
  boxrule=1.5pt,
  arc=3mm,
  boxsep=5pt,
  left=5pt,
  right=5pt
]
Transform this document into exactly \{target\_words\} words.
\end{tcolorbox}

\item \textbf{Thinking V1}: Our Plan-and-Write approach with explicit word counting.

\begin{tcolorbox}[
  title={\textbf{Thinking V1 Prompt}},
  colback=orange!10,
  colframe=orange!50,
  coltitle=black,
  boxrule=1.5pt,
  arc=3mm,
  boxsep=5pt,
  left=5pt,
  right=5pt
]
YOUR ONLY TASK: Summarize this document in EXACTLY \{target\_words\} WORDS.

I will only accept summaries with EXACTLY \{target\_words\} words - count carefully.

STEP 1: Count each word as you write:
\begin{verbatim}
<thinking>
1 First
2 word
...
{target_words} lastword
</thinking>
\end{verbatim}

STEP 2: Rewrite those SAME \{target\_words\} WORDS in paragraph form:
\begin{verbatim}
<final_answer>
First word... 
[EXACTLY {target_words} WORDS TOTAL]
</final_answer>
\end{verbatim}
\end{tcolorbox}

\item \textbf{Thinking V2}: A scientific framing of our approach with structured planning.

\begin{tcolorbox}[
  title={\textbf{Thinking V2 Prompt}},
  colback=purple!10,
  colframe=purple!50,
  coltitle=black,
  boxrule=1.5pt,
  arc=3mm,
  boxsep=5pt,
  left=5pt,
  right=5pt
]
TASK: Transform this document to EXACTLY \{target\_words\} words while maximizing information preservation.

SCIENTIFIC METHODOLOGY:
\begin{enumerate}
\item First, identify the core information hierarchy and key points
\item Then perform controlled expansion to EXACTLY \{target\_words\} words by:
  \begin{itemize}
  \item Preserving primary information structures
  \item Including supporting details proportionally
  \item Maintaining relative emphasis on topics from original document
  \item Adding clarifying context where needed to reach target length
  \end{itemize}
\end{enumerate}

EXECUTION:
\begin{verbatim}
<thinking>
- First outline core information structure
- Draft initial version
- Systematically expand by adding supporting 
examples
- Count words meticulously: 1, 2, 3... 
until reaching {target_words}
</thinking>

Final {target_words}-word document:
\end{verbatim}
\end{tcolorbox}

\end{enumerate}

\subsubsection{Generation Protocol}
For each model and target length combination, we conducted five independent attempts to account for generation variance. This resulted in 960 individual generations (6 models $\times$ 8 target lengths $\times$ 5 attempts $\times$ 4 prompt variants). Between attempts, we implemented a delay to manage API rate limits and ensure model availability.

\subsubsection{Word Counting Methodology}
We employed NLTK's word\_tokenize function to standardize word counting across all outputs, excluding punctuation marks. For the ``thinking'' prompt variants, we extracted only the final answer text from between the designated tags to evaluate word count accuracy.

\subsubsection{Analysis Framework}
After collecting results, we computed statistical measures including mean and standard deviation of word count accuracy across models and prompt variants. We also conducted significance testing to determine whether improvements from the Plan-and-Write methodology were statistically significant compared to baseline approaches.

%%
%% RESULTS
\section{Results}

Our evaluation reveals significant differences in length control capabilities across models and prompting strategies. We present results from document summarization task, examining how our Plan-and-Write approach compares to standard prompting methods.

\begin{figure*}
    \centering
    \includegraphics[width=\textwidth]{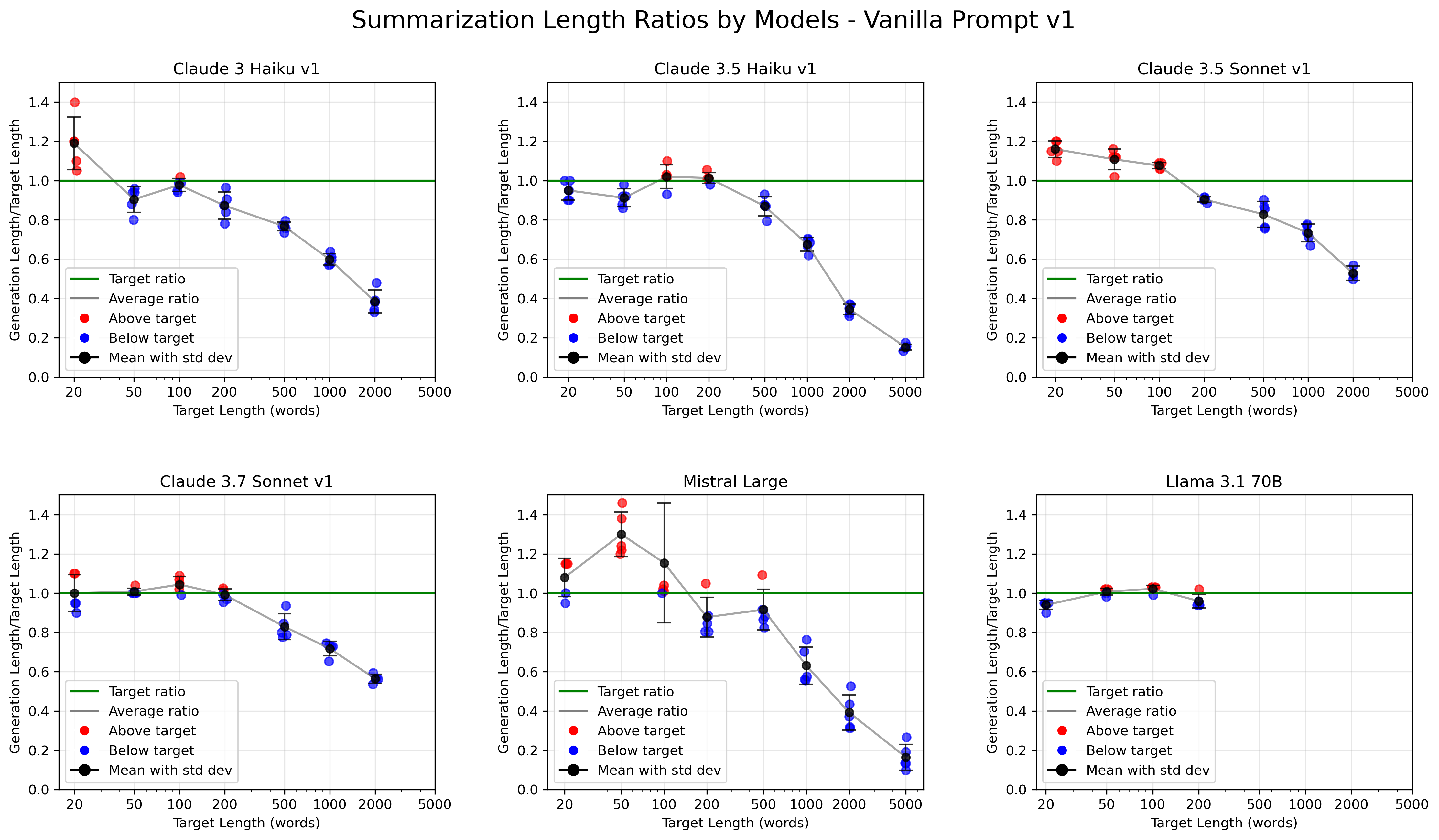}
    \caption{Length fidelity with vanilla prompting. The vertical axis shows the ratio of generated length to target length, with 1.0 representing perfect adherence. Each point represents an individual generation attempt, with color indicating over-generation or under-generation. Black points with error bars show the mean and standard deviation for each target length.}
    \label{fig:vanilla_results}
\end{figure*}

With vanilla prompting, most models exhibit considerable variation in output length, with a systematic bias toward over-generation, particularly for shorter targets. In contrast, our Plan-and-Write approaches demonstrate tighter clustering around the ideal ratio of 1.0 across all length targets, providing visual evidence that explicit word counting significantly improves length fidelity.

\begin{figure*}
    \centering
    \includegraphics[width=\textwidth]{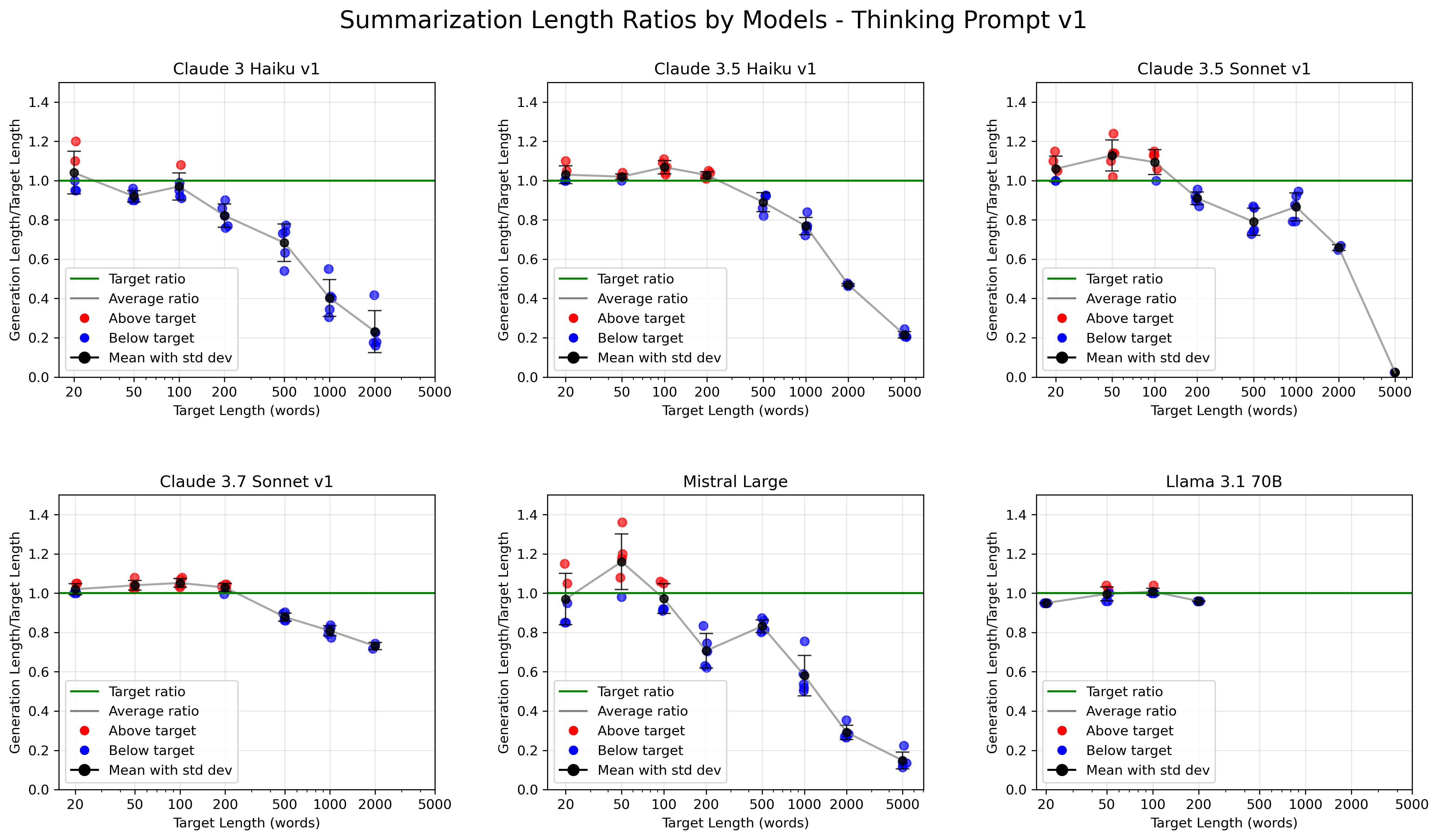}
    \caption{Length fidelity with Plan-and-Write prompting. Note the tighter clustering around the target ratio of 1.0 compared to vanilla prompting, indicating improved length control across most models.}
    \label{fig:thinking_results}
\end{figure*}

\begin{table*}
\caption{Mean Absolute Percentage Deviation (MAPD $\pm$ standard deviation) across models and prompting strategies for document summarization. Lower values indicate better length control. Best method for each model is highlighted in bold.}
\label{tab:mapd_results}
\centering
\begin{tabular}{@{}lccccc@{}}
\toprule
\textbf{Model} & \textbf{Vanilla V1} & \textbf{Vanilla V2} & \textbf{Thinking V1} & \textbf{Thinking V2} & \textbf{Best} \\
\midrule
Claude 3 Haiku & 0.242 ± 0.200 & 0.130 ± 0.083 & 0.297 ± 0.273 & \textbf{0.120 ± 0.097} & Thinking V2 \\
Claude 3.5 Haiku & 0.271 ± 0.301 & 0.330 ± 0.363 & \textbf{0.225 ± 0.270} & 0.316 ± 0.352 & Thinking V1 \\
Claude 3.5 Sonnet & 0.176 ± 0.119 & 0.308 ± 0.291 & \textbf{0.159 ± 0.172} & 0.192 ± 0.115 & Thinking V1 \\
Claude 3.7 Sonnet & 0.141 ± 0.146 & 0.268 ± 0.307 & \textbf{0.088 ± 0.079} & 0.177 ± 0.149 & Thinking V1 \\
Llama 3.1 70B & 0.037 ± 0.023 & \textbf{0.027 ± 0.019} & 0.032 ± 0.020 & 0.036 ± 0.042 & Vanilla V2 \\
Mistral Large & \textbf{0.328 ± 0.279} & 0.361 ± 0.274 & 0.349 ± 0.283 & 0.402 ± 0.262 & Vanilla V1 \\
\bottomrule
\end{tabular}
\end{table*}

Table~\ref{tab:mapd_results} presents the Mean Absolute Percentage Deviation (MAPD) for each model across four prompting strategies. Lower values indicate better performance. Several key findings emerge from our analysis:

\begin{itemize}
    \item \textbf{Plan-and-Write Effectiveness:} Our structure-guided approaches (Thinking V1 and V2) outperformed standard prompting for four out of six models, demonstrating their broad applicability across model architectures. For Claude 3.7 Sonnet, Thinking V1 reduced MAPD to 0.088, representing a 37.6\% improvement over the best vanilla approach. For Claude 3 Haiku, Thinking V2 achieved the best results with an MAPD of 0.120.

    \item \textbf{Model-Specific Performance:} Llama 3.1 70B demonstrated exceptional length control across all prompting strategies, achieving the lowest overall MAPD of 0.027 with the Vanilla V2 prompt. This suggests that some models may have already developed strong length control capabilities during their training.

    \item \textbf{Length Scaling:} As shown in Figures~\ref{fig:vanilla_results} and~\ref{fig:thinking_results}, most models demonstrated better length adherence for longer target lengths (500+ words) compared to shorter targets. This pattern was consistent across prompting strategies but was less pronounced with our Plan-and-Write approaches.

    \item \textbf{Variance Reduction:} The standard deviation of MAPD was typically lower for our thinking approaches compared to vanilla prompts, especially for more advanced models. This indicates that explicit word counting leads to more consistent and predictable length control.
\end{itemize}

Our results show a correlation between model capabilities and length control performance. More advanced models like Claude 3.7 Sonnet and Llama 3.1 70B demonstrated superior length fidelity across all prompting strategies. However, the relative improvement from structure-guided prompting was most pronounced in mid-tier models like Claude 3.5 Sonnet and Claude 3.5 Haiku, suggesting that explicit planning mechanisms are particularly beneficial for these models.

\subsection{Quality Evaluation with LLM-as-a-Judge}

While length fidelity is our primary focus, it is essential to ensure that improved length control does not come at the expense of output quality. To systematically evaluate content quality across prompting strategies, we employed LLM-as-a-Judge (LLMaaJ), an increasingly common evaluation method that leverages large language models to assess semantic properties beyond surface-level metrics.

We assessed four key dimensions of quality across all model outputs:

\begin{itemize}
    \item \textbf{Correctness}: Measures the factual accuracy of information presented in the summary relative to the source document, evaluating whether statements accurately reflect information from the original text.

    \item \textbf{Completeness}: Assesses whether the summary captures all essential information from the original document proportionate to its length target, including key points, arguments, and conclusions.

    \item \textbf{Faithfulness}: Evaluates whether the summary contains information that is consistent with the source document without introducing facts or claims not present in the original.

    \item \textbf{Relevance}: Measures how well the summary focuses on information that matters to the core message of the document, avoiding tangential details while highlighting central themes.
\end{itemize}

\begin{table*}
\caption{Quality evaluation results across different prompting strategies using LLM-as-a-Judge. Higher scores indicate better performance (0-1 scale). Best method for each quality dimension is highlighted in bold.}
\label{tab:quality_results}
\centering
\begin{tabular}{lcccc}
\hline
\textbf{Prompting Strategy} & \textbf{Correctness} & \textbf{Faithfulness} & \textbf{Completeness} & \textbf{Relevance} \\
\hline
Vanilla V1 & \textbf{0.91} & 0.95 & 0.81 & 0.71 \\
Vanilla V2 & \textbf{0.91} & 0.93 & 0.73 & 0.69 \\
Thinking V1 & 0.90 & \textbf{0.96} & \textbf{0.85} & \textbf{0.87} \\
Thinking V2 & 0.87 & 0.94 & 0.84 & 0.73 \\
\hline
\end{tabular}
\end{table*}

Table~\ref{tab:quality_results} presents the quality evaluation results across all prompting strategies. Notably, our Plan-and-Write approach (Thinking V1) not only improved length fidelity but also enhanced summary quality in several dimensions. While Vanilla approaches scored marginally higher on correctness (0.91 vs.~0.90), Thinking V1 achieved the highest scores in faithfulness (0.96), completeness (0.85), and relevance (0.87). These results suggest that structure-guided prompting can simultaneously improve length adherence and enhance output quality.

\subsection{Open-Weight Model Evaluation}

To evaluate generalizability across model families, we conducted additional experiments with the open-weight Qwen 2.5 7B model deployed on AWS SageMaker, following identical experimental protocols as our primary evaluation. Our findings revealed that the Thinking V2 approach achieved the lowest overall MAPD (0.280 ± 0.636), marginally outperforming Vanilla V2 (0.281 ± 0.169).

\begin{figure*}
    \centering
    \includegraphics[width=\textwidth]{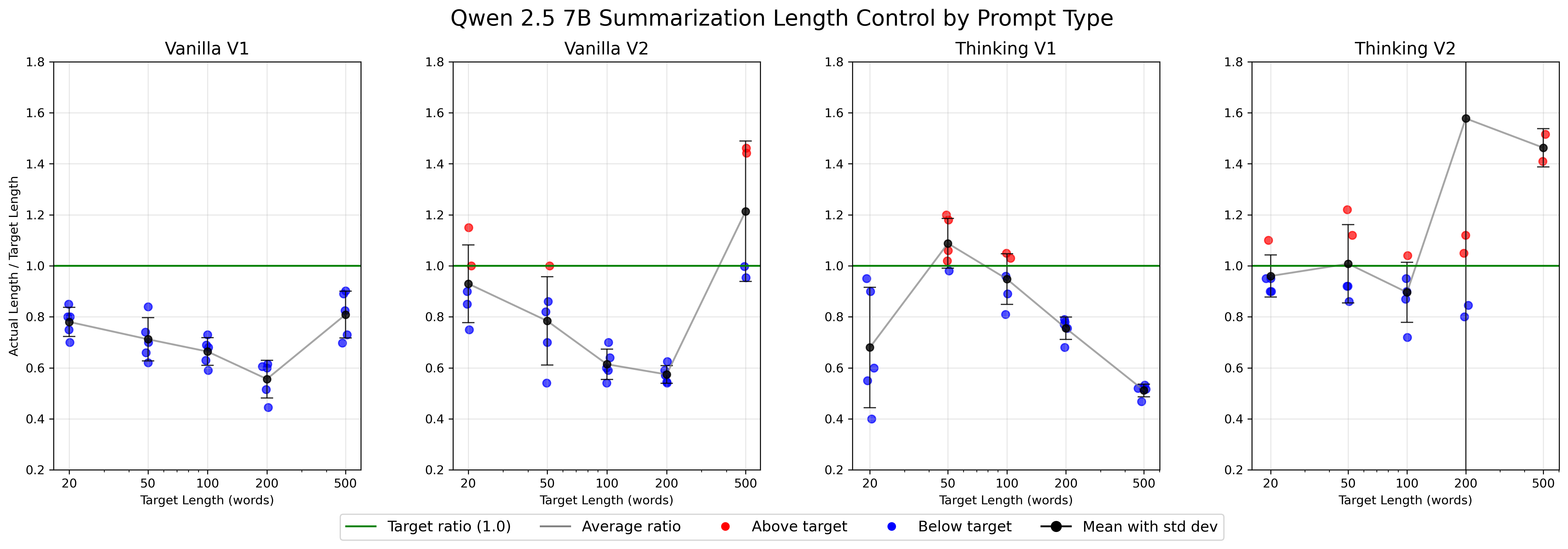}
    \caption{Length fidelity comparison across four prompting strategies for Qwen 2.5 7B.}
    \label{fig:qwen_results}
\end{figure*}

\subsection{Trade-off Analysis}

While structure-guided prompting improves length fidelity for several models, production deployments must consider computational trade-offs. We conducted a detailed cost-benefit analysis using Qwen 2.5 7B deployed on an AWS ml.g5.12xlarge instance, measuring both token consumption and inference latency:

\begin{table}[h]
\caption{Cost-benefit analysis of different prompting strategies with Qwen 2.5 7B on AWS.}
\label{tab:cost_analysis}
\centering
\begin{tabular}{lcc}
\toprule
\textbf{Metric} & \textbf{Vanilla} & \textbf{Thinking} \\
\midrule
Average tokens & 7,914 & 8,046 (1.02×) \\
Average latency & 1,000.1 ms & 1,573.8 ms (1.57×) \\
\bottomrule
\end{tabular}
\end{table}

Our analysis reveals that thinking prompts require only marginally more tokens (1.02×) but substantially longer inference time (1.57×) compared to vanilla prompts. For this specific model, the increased computational cost did not translate to improved length fidelity, suggesting that simpler prompting strategies may be more cost-effective in this case.

%%
%% LIMITATIONS
\section{Limitations}

While our Plan-and-Write methodology shows promising results, several limitations warrant acknowledgment. First, the approach does not benefit all models equally---some models like Llama 3.1 70B already exhibit strong length control capabilities with standard prompting, limiting the relative improvement from our method. Second, we observed that for longer target lengths (beyond 500 words), length fidelity tends to decrease across all prompting strategies, suggesting fundamental limitations in LLMs' ability to maintain precise counting over extended outputs. Third, the two-phase generation process introduces additional computational overhead by requiring models to generate more tokens during the planning phase, potentially increasing inference costs. Finally, while we observed no significant quality degradation in model outputs, our quality evaluation relies on LLM-as-a-Judge, which may introduce its own biases.

%%
%% CONCLUSION
\section{Conclusion}

This paper introduces Plan-and-Write, a prompt engineering methodology for precise length control in large language models without requiring model retraining. Through comprehensive evaluation across seven state-of-the-art LLMs, we demonstrated that explicit word counting and structured planning can significantly improve length fidelity compared to standard prompting techniques. Our results show that five of seven tested models benefit from our approach, with improvements in Mean Absolute Percentage Deviation of up to 37.6\%. Our quality evaluation using LLM-as-a-Judge demonstrates that the methodology not only improves length control but also maintains or enhances overall output quality.

Our work offers immediate practical value for developers and researchers working with black-box LLMs in production environments where model retraining is impractical or cost-prohibitive. The Plan-and-Write methodology provides a straightforward, deployable solution for applications requiring precise length control, from voice interfaces to mobile applications with display constraints. Future research directions include exploring hybrid approaches that combine prompt engineering with lightweight inference-time modifications, investigating the generalizability of structure-guided prompting to other types of constraints beyond length, and developing adaptive prompting strategies that adjust based on model capabilities and specific length targets.

%%
%% Bibliography
\newpage
\bibliographystyle{ACM-Reference-Format}
\bibliography{sample-base}

%%
%% APPENDIX
\onecolumn
\appendix

\section{Length Fidelity Plots for Summarization}

\begin{figure*}[!ht]
    \centering
    \includegraphics[width=0.88\textwidth]{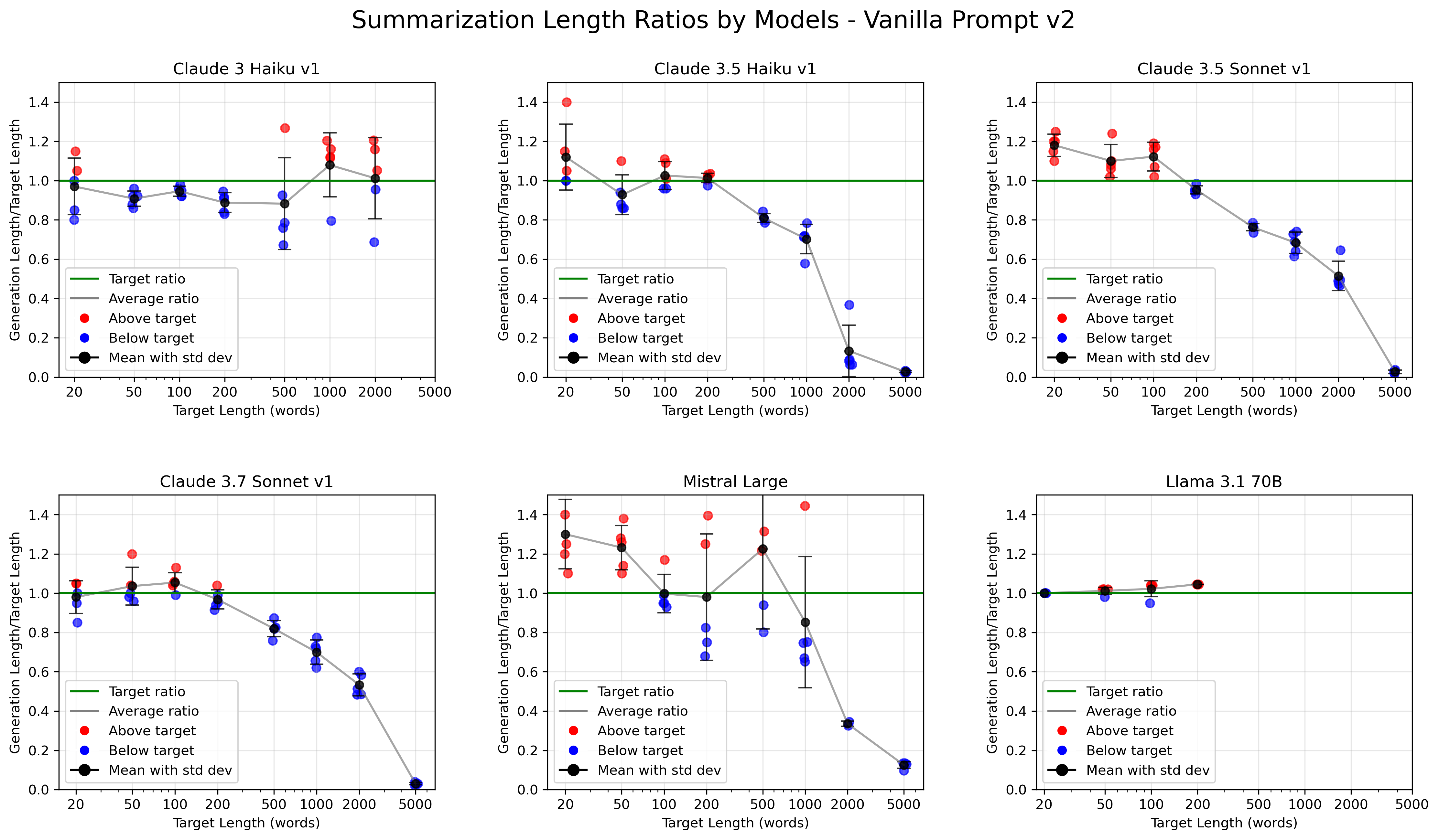}
    \caption{Length fidelity with vanilla prompting v2.}
    \label{fig:vanilla_v2_results}
\end{figure*}

\begin{figure*}[!ht]
    \centering
    \includegraphics[width=0.88\textwidth]{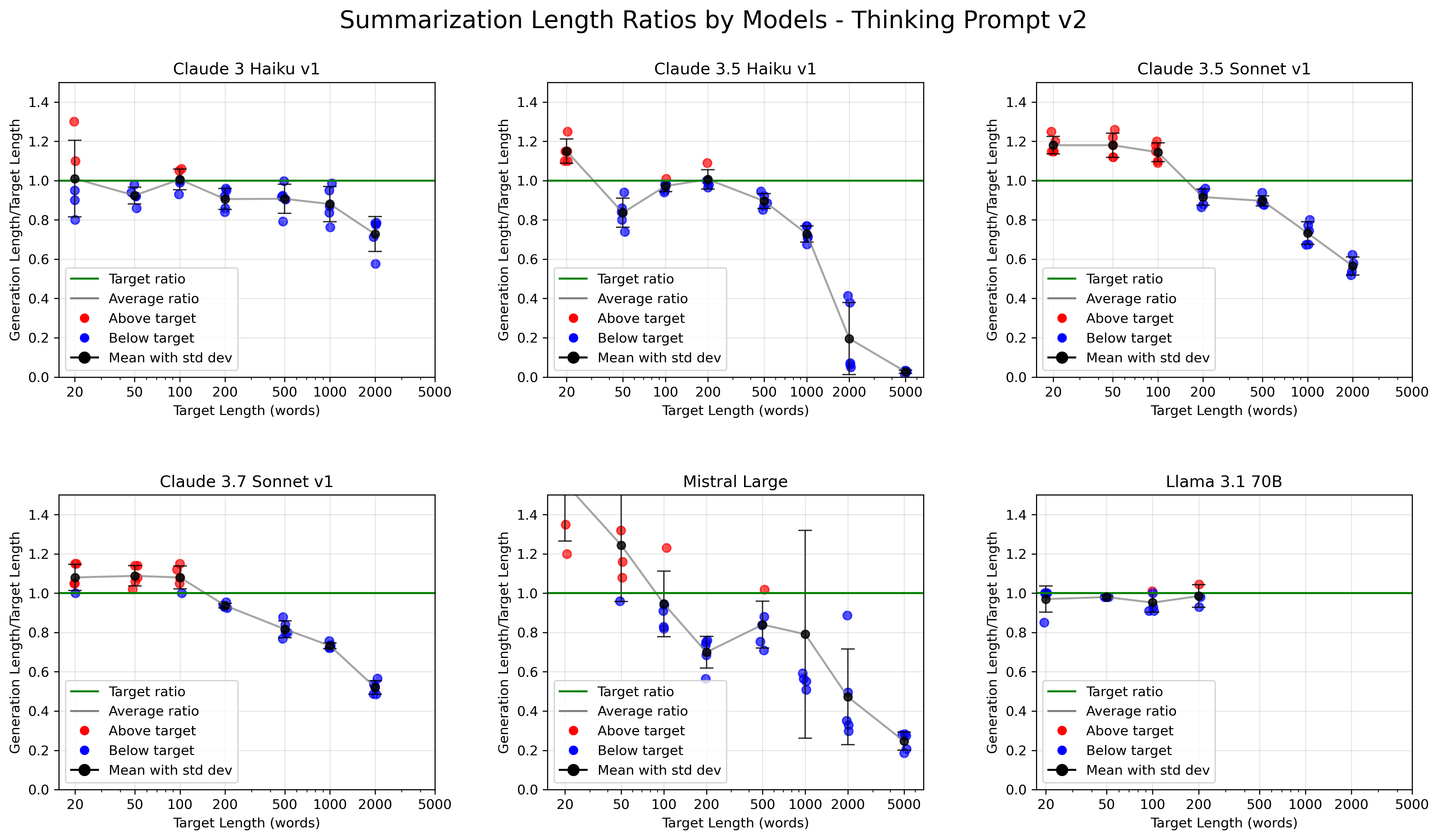}
    \caption{Length fidelity with Plan-and-Write prompting v2.}
    \label{fig:thinking_v2_results}
\end{figure*}

\newpage

\section{Story Generation Length Fidelity}

In addition to document summarization, we conducted experiments on creative story generation tasks to evaluate whether our structure-guided Plan-and-Write approach would yield similar benefits for open-ended content creation. We prompted models to write stories about a young boy discovering a magical book, providing specific target word counts.

\begin{tcolorbox}[
  title={\textbf{Vanilla Prompt for Story Generation}},
  colback=blue!10,
  colframe=blue!50,
  coltitle=black,
  boxrule=1.5pt,
  arc=3mm,
  boxsep=5pt,
  left=5pt,
  right=5pt
]
Write a story about a young boy who discovers a magical book in his attic and learns how to harness the power of magic within himself in exactly \{target\_words\} words.
\end{tcolorbox}

\vspace{1em}

\begin{tcolorbox}[
  title={\textbf{Thinking Prompt for Story Generation}},
  colback=orange!10,
  colframe=orange!50,
  coltitle=black,
  boxrule=1.5pt,
  arc=3mm,
  boxsep=5pt,
  left=5pt,
  right=5pt
]
TASK: Write a story about a young boy who discovers a magical book in his attic and learns how to harness the power of magic within himself using EXACTLY \{target\_words\} words.

SCIENTIFIC METHODOLOGY:
1. First, outline the key story elements and narrative arc
2. Then perform controlled story development to EXACTLY \{target\_words\} words by:
   a) Establishing setting, characters, and conflict
   b) Developing plot points proportionally
   c) Maintaining narrative coherence and flow
   d) Including appropriate details to reach target length

CONSTRAINTS:
- Output MUST contain EXACTLY \{target\_words\} words
- Story should be engaging and complete
- Narrative complexity should scale with target length
\end{tcolorbox}

\subsection{Story Generation Results}

Our results for story generation tasks reveal a notable contrast with document summarization findings. The results demonstrate that Plan-and-Write did not provide consistent benefits for creative generation tasks---in fact, for five of six models, vanilla prompting achieved better length fidelity. This suggests that structure-guided prompting may be more effective for information condensation than for creative content generation.

\begin{table}[h]
\caption{Mean Absolute Percentage Deviation (MAPD $\pm$ std dev) for story generation. Lower values indicate better length control.}
\label{tab:story_results}
\centering
\begin{tabular}{@{}lccc@{}}
\toprule
\textbf{Model} & \textbf{Vanilla MAPD} & \textbf{Thinking MAPD} & \textbf{Best} \\
\midrule
Claude 3 Haiku & \textbf{0.080 ± 0.093} & 0.113 ± 0.082 & Vanilla \\
Claude 3.5 Haiku & \textbf{0.204 ± 0.220} & 0.844 ± 2.015 & Vanilla \\
Claude 3.5 Sonnet & \textbf{0.045 ± 0.031} & 0.084 ± 0.054 & Vanilla \\
Claude 3.7 Sonnet & \textbf{0.047 ± 0.030} & 0.061 ± 0.042 & Vanilla \\
Llama 3.1 70B & 0.133 ± 0.184 & \textbf{0.117 ± 0.191} & Thinking \\
Mistral Large & \textbf{0.266 ± 0.229} & 0.322 ± 0.286 & Vanilla \\
\bottomrule
\end{tabular}
\end{table}

\begin{figure}[h]
    \centering
    \includegraphics[width=0.88\columnwidth]{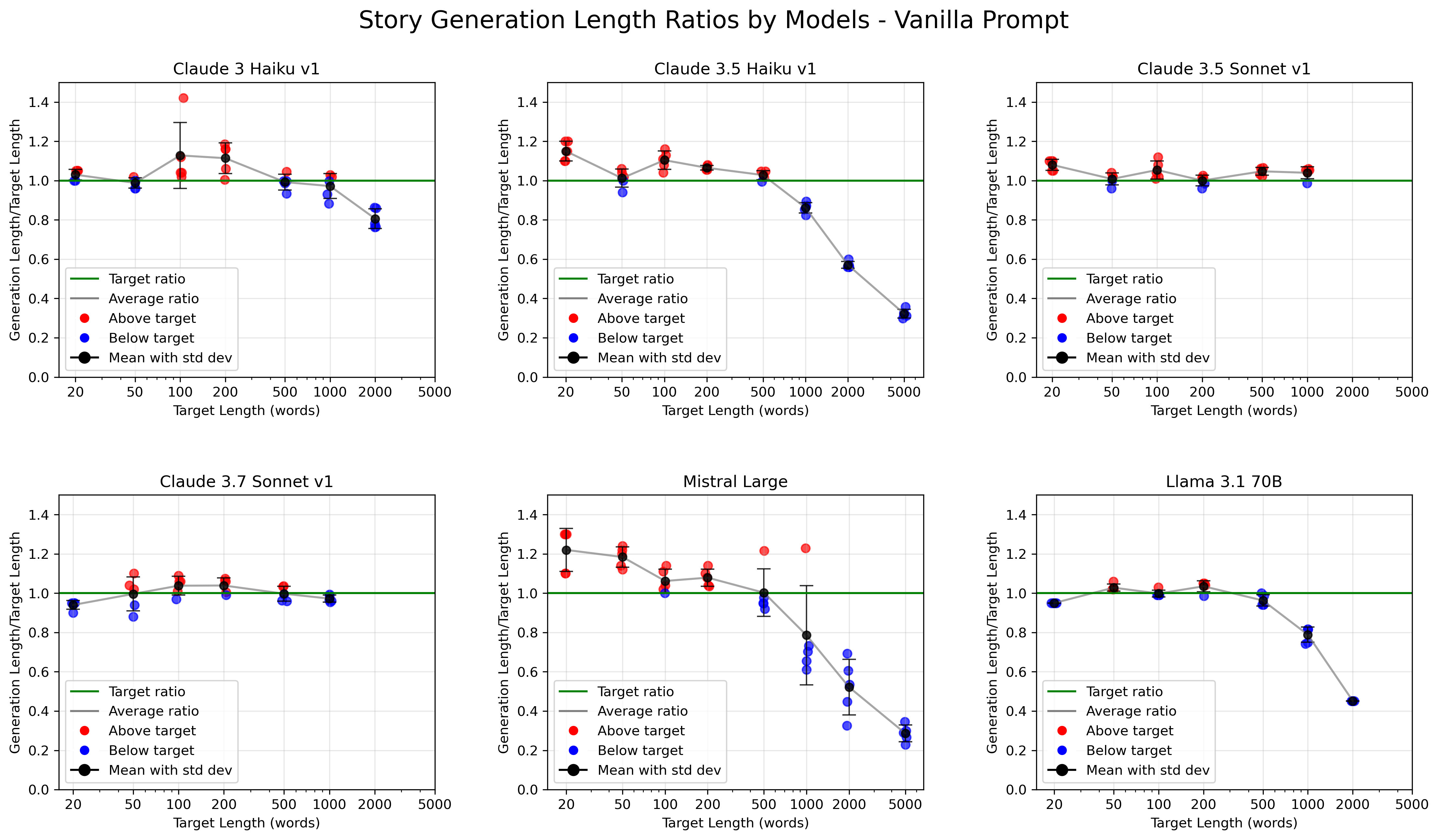}
    \caption{Length fidelity with vanilla prompting for story generation.}
    \label{fig:gen_vanilla_results}
\end{figure}

\begin{figure}[h]
    \centering
    \includegraphics[width=0.88\columnwidth]{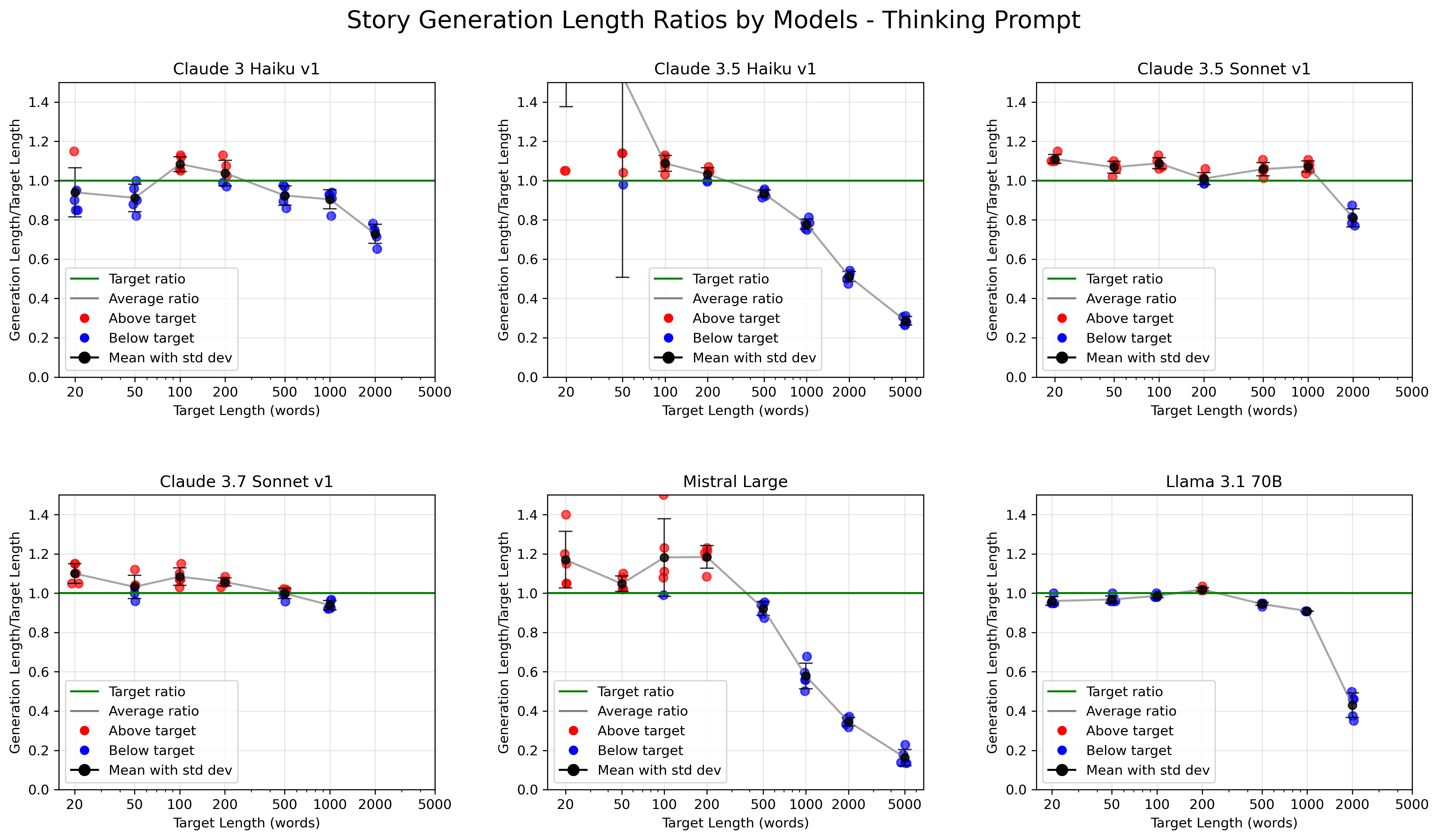}
    \caption{Length fidelity with Plan-and-Write prompting for story generation.}
    \label{fig:gen_thinking_results}
\end{figure}

These contrasting results between summarization and story generation suggest that the benefits of structure-guided prompting may be task-dependent. Specifically, explicit planning and word counting appear more beneficial for tasks that require condensing existing information (like summarization) than for tasks requiring creative content generation.

\end{document}